# Machine Learning
# For Distributed Acoustic Sensors
# Classic versus Image and Deep Neural Networks Approach

Mugdim Bublin
FH Kaernten

**ABSTRACT**

Distributed Acoustic Sensing (DAS) using fiber optic cables is a promising new technology for pipeline monitoring and protection. In this work, we applied and compared two approaches for event detection using DAS: Classic machine learning approach and the approach based on image processing and deep learning. Although with both approaches acceptable performance can be achieved, the preliminary results show that image based deep learning is more promising approach, offering six times lower event detection delay and twelve times lower execution time.

## 1. INTRODUCTION

At present, pipeline monitoring and protection is performed by expensive and laborious methods like helicopter or vehicle patrols. Distributed acoustic-optical sensors (DAS) are a comparatively new approach (see Figure 1. ). Such systems permit supervision of long distance pipelines using fiber optics. Since fiber optic cables are usually laid along the modern pipelines no additional investment in equipment installation is needed.

But the application of distributes acoustic-optical sensors requires advanced computer algorithms capable of recognizing events of interest like manual and/or excavator digging, tapping and intrusion detection. Finding efficient algorithms for event detection based on DAS measurements is especially hard due to huge variety of scenarios like various kinds of excavators and soil types, different temperature and weather conditions, as well as possible interference by other signal sources like highways, railways, wind turbines, agricultural machines etc. Research and development of such algorithms is still in its infancy.

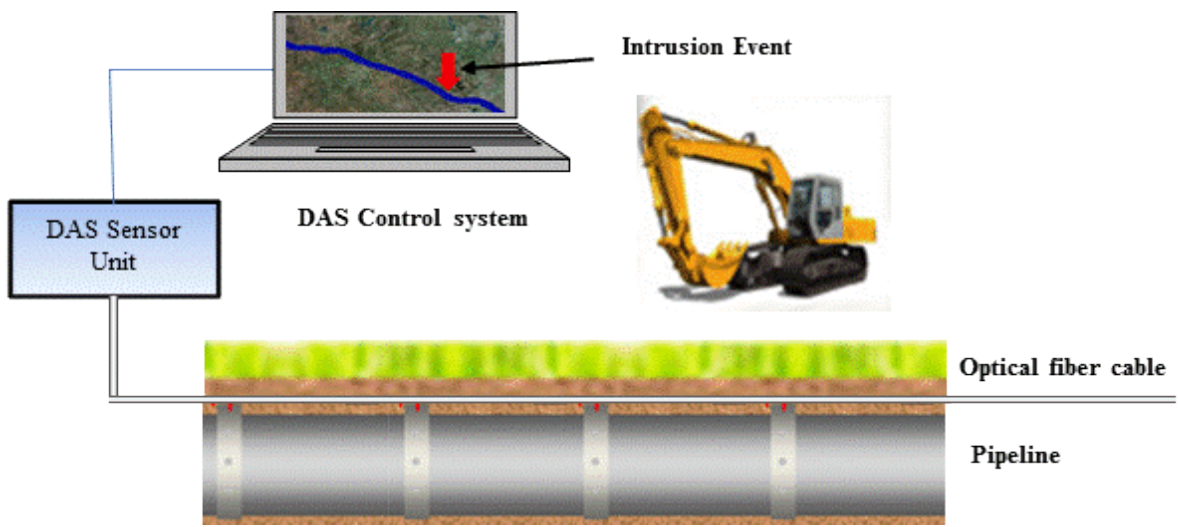

Figure 1. Distributed acoustic-optical sensors for pipeline monitoring

In this work, we present preliminary results of application of machine learning and image processing algorithms for event detection based on DAS signals generating along a pipeline. We implemented and compared two approaches for event detection using DAS: Classic machine learning approach and the approach based on image processing and deep learning.

In the sequel of the work we provide a short description of DAS technology. Then we describe some machine learning and image processing algorithms that can be used for event detection using DAS signals. Finally, we present and discuss results and provide suggestions for further work.

## 2. DISTRIBUTED ACOUSTO-OPTICAL SENSORS

Optical Time-Domain Reflectometry (OTDR) is a well-established technique used to check long-haul fibre optical connections in the telecommunication domain [1]. This technology is based on emitting short pulsed into the fiber, and recording the intensity of light reflected to the sender by Rayleigh reflection.

A distributed acousto-optical sensor can be constructing by exploiting the fact that the refractive index of a glass fiber is slightly affected by any applied pressure – including sound pressure: Short pulses are emitted as for OTDR, but instead of analyzing the intensity one evaluates the phase the Rayleigh reflected optical signal.

The phase of an optical signal is measured by using an interferometer and a delay line which brings light reflected at different distances – say, 10 m apart – to the interferometer at the same time (see Figure 2. ). Alternatively, no delay line is needed if the pulse forming unit may create two short pulses emitted on in the fiber at say 20m.

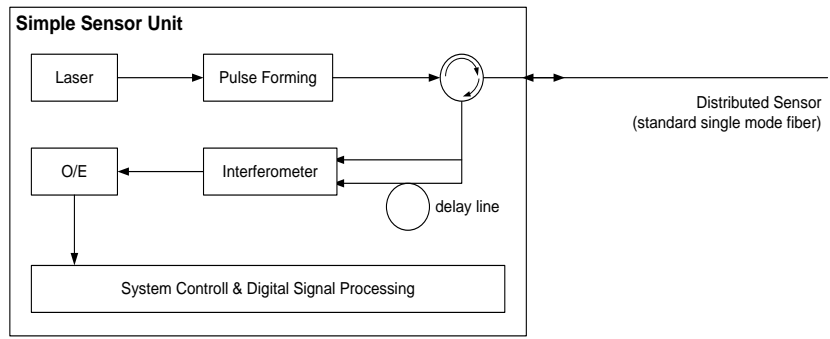

Figure 2.   Distributed Acoustic Sensor (DAS) Unit

By evaluating the interferometer every 100 nanoseconds this procedure captures the sound pressure at every ~10m. By emitting pulses at a rate of say 1000 Hz one obtains a distributed acoustic-optical sensor which is capable of detect the sound pressure up to distances of 40 km at regular distance of 10 m and up to acoustic frequencies of 500 Hz.

Optical fiber sensors have certain advantages that include immunity to electromagnetic interference, lightweight, small size, high sensitivity, large bandwidth, and ease in implementation as fiber cables are often already installed for communication purposes in critical infrastructures. Strain, temperature and pressure are the most widely studied measurables used in optical fiber sensing [2-6]. The tiny pressure change induced by acoustic events in the surrounding of an optical fiber can be measured by optical means over large distance and therefore surveillance of large areas becomes possible using this technology.

Most of the current investigation and available literature focuses on the extraction of the signal from the fiber which is done by interferometer assemblies. In [1] the sensing of vibration using an optical frequency-domain reflectometry (OFDR) technique has been demonstrated in a 17 m length fiber up to a frequency of 32 Hz. A Mach-Zender interferometer has been used together with standard single-mode fibers. However, the work focuses of the extraction of the signal and the classification of events from the signal is not discussed.

Choi [7] reports on the use of OTDR for intruder detection over a cable buried 30 cm deep in the ground and with a length of 2 km, tested with a 60 kg person walking in up to 1 m distance from the cable. Again, any change in the signal was interpreted as evidence for the intrusion event without discussing the possibility to classify events by careful analysis of the signal signature. Kumagai [8] describes an approach to distinguish intruders climbing a fence from fence vibration due to wind to minimize false alarms. The signal of a Sagnac type interferometer was used to classify the event using FFT frequency analysis in the 0 to 250 Hz range. It has been demonstrated that the event of a person climbing the fence can be distinguished from the wind induced fence vibration and the resulting false alarm rate was around 1 per month during a 1 year test period. About 2 real events per day have been detected with a 100% detection rate.

Juarez et al. [2] describe the OTDR system for the detection of intruders in laboratory and field tests using a 12 km length fiber. In the field tests has been successfully demonstrated ability to detect an 80 kg person walking over the fiber. But the ability to classify different events has not been discussed nor investigated.

US patent 5,194,847 [9] discloses a method and apparatus for intrusion detection based on the OTDR technique. Details of optical setup and pulse forming as well as signal processing for the detection and location of the intruder from the backscattered signal of an interrogating pulse are described, but the classification of the event is not disclosed.

Harman discloses a method and apparatus in PCT patent application WO2013/185208 [10] for short range perimeter surveillance with two back-to-back Michelson interferometers using a cable comprising four optical fibers. He targets to achieve a competitive price for a security installation using this arrangement and describes the necessary signal processing and post-processing techniques to extract an intruder's location from the optical signal. Again, signal processing for the classification of the event are not disclosed.

## 3. MACHINE LEARNING FOR DAS EVENT DETECTION

While the state of the art literature describes basic signal processing to resolve an event from the OTDR signal and thereby generate alarms, the classification of event types from the signal is in its infancy. There is a need to monitor and detect specific, safety relevant events with a low false alarm rate among a larger number of "harmless" background events. The state of the art is insufficient for a successful deployment of the OTDR system for robust road security surveillance and pipeline intrusion detection.

Although it is a new research area requiring multidisciplinary approach, we are standing on the "shoulders of giants" and can exploit a vast body of knowledge in signal processing and pattern recognition in general [11-13] and more recent advances like deep learning in special [14, 15].

The task of event recognition using the fiber optics, consists of signal (pre-) processing, pattern recognition (which itself consists of feature extraction and classification) and event tracking. For each of this task there are a plenty of algorithms proposed in classical ML literature (see Figure 3.).

The art of science of machine learning consists of selecting of appropriate features and classification algorithms for the task at hand. Especially feature selection is demanding task in classical machine learning that often requires high effort and very good domain knowledge. The big advantage of deep neural networks is their capability to extract the relevant features from the raw data in a hierarchical manner without need for much domain knowledge. One also doesn't need to search for optimal classifier for the selected features, since the deep neural networks offers a unified approach for feature selection and classification. These two advantages together with their superior performance make deep neural networks a promising approach for many machine learning tasks.

In following we provide a brief overview of the main algorithms used in machine learning that might be useful for our DAS event detection.

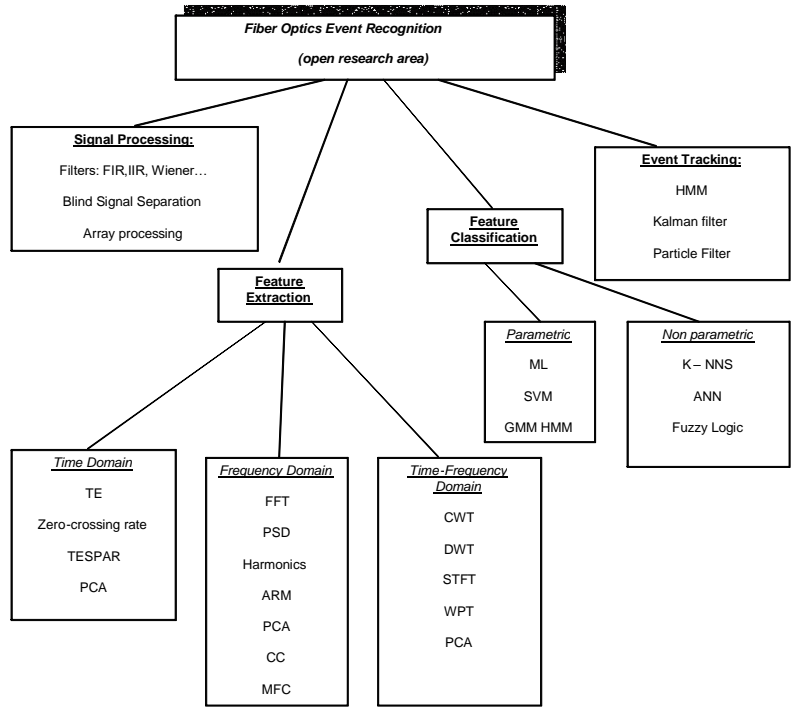

| ANN | Artificial Neural Network | k-NNS | k - Nearest Neighbor Search |
|---|---|---|---|
| ARM | Autoregressive Modeling | MFC | Mel-frequency cepstrum |
| CC | Cepstral coefficients | ML | Maximum-Likelihood Method |
| CWT | Continuous Wavelet Transform | PCA | Principal Component Analysis |
| DWT | Discrete Wavelet Transform | PSD | Power Spectral Density |
| FFT | Fast Fourier Transformation | STFT | Short Time Fourier Analysis |
| FIR | Finite Impulse Response | SVM | Support vector machine |
| GMM | Gaussian Mixture Model | TE | Time Energy Distribution |
| HMM | Hidden Markov Model | TESPAR | Time Encoded Signal Processing and Recognition |
| IIR | Infinite Impulse Response Filter | WPT | Wavelet Packet Transform |

Figure 3. Overview of the classical algorithms for feature selection and machine learning

### 3.1 Feature Extraction

Before feature extraction, standard signal processing techniques like FFT, DWT, CWT, STFT, FIR, IIR and Wiener Filtering are used for data pre-processing and visualization, which are necessarily prerequisite for further signal analysis and pattern recognition [11]. The task of the feature extraction is to define and calculate the specific characteristics of the signals (features) that can be used for classification [12 – 13]. The outputs of feature extractions are N dimensional vectors that are mapped by feature classifier to the classes in a N dimensional feature space. The features can be extracted from signal characteristics in time, frequency or in time and frequency (scale) domain.

Following Features are often used in classical ML approach:

**Time Domain Features** like Energy distribution of the signal (TE), where the energy of a short time window of the source signal is used to discriminate between classes and Zero-crossing rate estimated by counting the number of zero crossings of a signal within a time interval.

**Principal Component Analysis (PCA)** is popular statistical tools that for dimensional reduction based on finding the principal eigenvectors of the signals covariance matrix. PCA can be used as a feature extraction method in time, frequency or time-frequency domain [12 – 13].

**Frequency based feature** generation methods are often used for signal classification whereby Fast Fourier Transform (FFT) and Power Spectral Density (PSD) are used to extract feature vectors.

Patterns in spectral peaks (harmonics) can be also used as spectral features. For example, the magnitudes of the harmonic frequency components are considered as the feature vectors.

Cepstral coefficients (CC) are the coefficients of the inverse Fourier Transform of the log of the magnitude of the spectrum [13]. Mel-frequency cepstrum (MFC) is a representation of the short-term power spectrum of a sound, where the log power spectrum on a nonlinear mel scale of frequency is transformed based on a linear cosine transform.

**Time-Frequency Domain Features**: Short Time Fourier Transform (STFT) can be used to transform the overlapped acoustic Hamming windowed frames to a feature vector [13].

Also, **wavelet** analysis methods are suitable for the extraction of signal features [13].

### 3.2 Feature Classification

The task of feature classifiers is to assign the N dimensional feature vectors obtained by feature extraction to the different classes. In our case there are two classes: Intrusion (Excavator) and no intrusion. In general, it means dividing N dimensional feature space in several regions and associating the regions to the classes. In following we provide an overview of some popular classifiers according to [12 – 15].

**Support Vector Machines (SVM)** separates the classes by choosing the hyperplane that maximize the distance between the hyperplane and the closest points in each feature space region which are called support vectors. For the cases that feature vectors are nonlinear separable, a kernel function maps the input vectors to a higher dimension space in which a linear hyperplane can be used to separate the vectors.

**Decision trees** are (tree-like) graphs in which each internal node represents an "if"-test on a feature, each branch represents the outcome of the test, and each leaf node represents a class label.

**Pruned Trees** are reduced size decision trees where some sections are eliminated in order to increase generalization capabilities and avoid overfitting.

**Artificial Neural Networks (ANN)** are classifier methods that don't need special assumptions on the underlying probability models [12 – 15]. A popular version of the ANN are so called **feed forward** ANN that take features as an input layer and forward signals from the neurons of one hidden layer to the other hidden layer and finally to the output layer which usually provides the class probabilities. ANN can learn from examples i.e. adapt internal weights between neurons so that the predefined feature vectors are in certain sense optimally allocated to the predefined classes (learning examples).

Among ANN algorithms especially **Deep Neural Networks** with many layers of neurons organized in hierarchical manner drawn attention in recent time, due to their superb classification performance and their ability to extract features from raw data [14 -15]. As input signal for event detection an image can be used, that is previously generated from sensor measurements by diverse image processing algorithms [18-19]. After that Deep Neural Networks can operate on images to find the DAS events. For image classification, especially effective are **Convolutional Neural Networks (CNN)** [14] that hierarchically extract features from an image. For example, edges in images are extracted by a first hidden layer, than shapes are extracted in the second layers using edges from the first layer as the input, and finally the whole objects are extracted in the last hidden layer.

### 3.3 Tracking and Probability Evaluation

After detecting and classifying events additional algorithms could be used for tracking the events over time and space as well as evaluating the total probability of the events.

**Event Tracking**: The positions of events like single vehicles or group of vehicles can be tracked over time using well known filtering algorithms like HMM, Kalman or particle filtering [16 -17]. Using information about past positions and current measurements to determine the actual positions in general provides higher event detection probability than using the actual measurements alone.

For final evaluating event probabilities estimated by different classifiers and/or tracking algorithms described above, **Bayesian (Network) approach** can be used [20-21].

## 4. RESULTS AND DISCUSSION

We implemented and evaluated two approaches for event detection (see Figure 4. ): Classic machine learning approach and the approach based on image processing and deep learning.

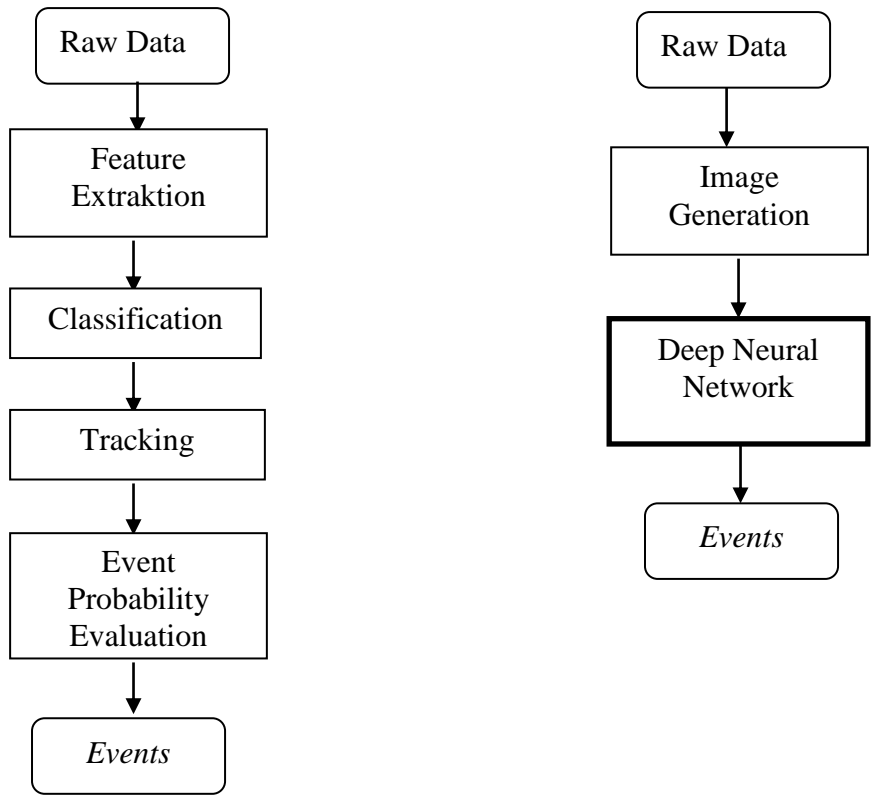

Figure 4.   Two methods for Event Detection by DAS: Classic machine learning approach (left) and Image Processing and Deep Neural Networks aproach (right)

Implementation of all algorithms for event detection is done in MATLAB using Signal Processing, Statistics and Machine Learning and Neural Networks Toolboxes. Implementation of basic signal processing operations that produces raw data from DAS measurements and maps the signals to Images is performed in C/C++ using Qt environment.

In following we describe two approaches: "Classic" machine learning approach and Deep Neural Networks approach using images in more details.

### 4.1 Classic Machine Learning Approach

"Classic" machine learning approach is based on Feature Extraction and Classification. In following we describe an application of the classic ML approach to the task of event detection with DAS systems.

**Feature Extraction**

The task of Feature Extraction is to extract from the raw data the features that are representative for the events of interest with enough discriminative power for subsequent Classification. As discussed in chapter 3 there are several possible features that can be used for event classification. We tried different features and feature combinations: Time, Frequency, Wavelets and Time/Frequency features. Among them the Frequency features turned out to be most appropriate features due their physical meaning and discriminative power. The first 100 FFT coefficients of the 1 second raw data signal sampled with 2kHz sample rate were used as classification features. It is important to stress that all signal characteristics are statistically distributed with relatively large variance, since signal changes depending of the source type and weight, distance, soil type, temperature etc. For example, Figure 5. represents spectrum statistics from an excavator signal and a highway signal obtained using about 100 sample signals for each event.

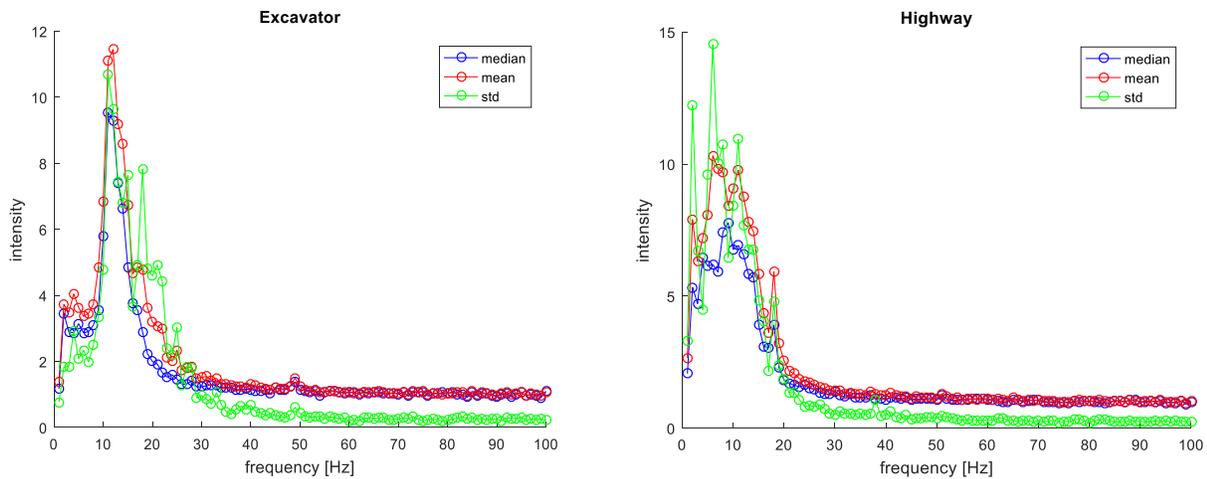

Figure 5. Spectrum statistics of an Excavator (left) and a Highway (right)

Note relatively high variations (standard deviation "std" in comparison to mean) in Figure 5. . The high variations in input features like spectrum make reliable classification i.e. distinguishing excavator signals from other signals like highway signals difficult.

**Classification**

The following classification algorithms were tested using 150000 signal samples (40000 Excavator signals and 110000 other signals): Support Vector Machine (polynomial order 3), Decision Trees, Pruned Tree and fully connected feed-forward Neural Networks with 5 hidden units.

In Table 1correct classification rates of different machine learning algorithms are presented.

**Table 1: Performance of different machine learning algorithms**

| ML Algorithm | Correct classifications [%] |
|---|---|
| SVM | 99,86 |
| Tree | 99,79 |
| Pruned Tree | 99,69 |
| Feedforward NN | 98,09 |

Although the classifier performance according to Table 1 may appear high, it should be considered that it is only for the signals of 1 second duration from one virtual sensor of the length of about 4m. Since a typical DAS unit covers the distance of about 40 km (about 10 000 virtual sensors) and a false alarm rate should be lower than once per month, we have a typical problem of rare event detection that often results in a high false alarm rate.

In order to keep false alarm rate low, we need also Tracking algorithms as denoted in Figure 4. (left) that keeps record of the positions and time steps of the detected events. Consecutive detections of the same event type in close positions are considered as a track of an event. For example, an excavator works relatively long (for a few minutes) at approximately (+/- 5m) same position.

Finally, event probability is evaluated taking into account classification probability of single detections and the track length in seconds. The higher the track length the more probable the event is. It turned out that for achieving high excavator detection reliability (99,9% at the distance below 10m) and low enough false alarm rate (maximum one false alarm per month) one needs 90 seconds i.e. 90 single excavator detections (one per second) at approximately same position.

The question is, if the detection delay of 90 seconds can be farther reduced by some other approaches to faster prevent possible intrusions. Furthermore, with above described "classic" machine learning approach the execution time is one second for evaluating one second of real time data over the length of 17 km i.e. the algorithms limit is about 17 km. Increasing the DAS unit range would directly decrease deployment and maintenance costs. It turns out that both: reduction of detection delay and the execution time is possible using Image Processing and Deep Neural Networks approach.

### 4.2 Image Processing and Deep Neural Networks Approach

Our Image Processing and Deep Neural Networks approach is based on the idea of converting the sensor signals first to (grey) image and then applying a deep neural network to recognize the events in the image.

To produce image from raw sensor data, at first root-mean-square (RMS) of 10 ms data (100 samples) for each virtual sensor (4 m) is calculated. The low-pass filtered RMS values are then processed by a horizontal and vertical Sobel filter to produce at the end a grey image (see example in Figure 6.

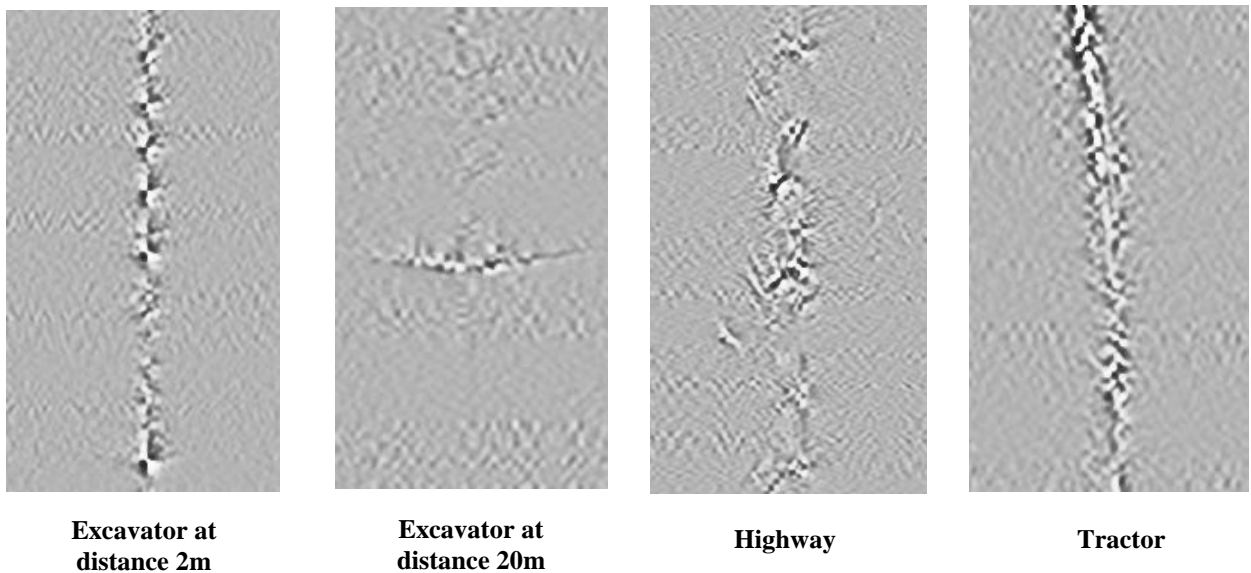

**Excavator at distance 2m**  **Excavator at distance 20m**  **Highway**  **Tractor**

Figure 6. Images of different signal sources

In the next step a deep neural network is trained using 40000 image samples to classify excavator versus not excavator events (see Figure 7. ).

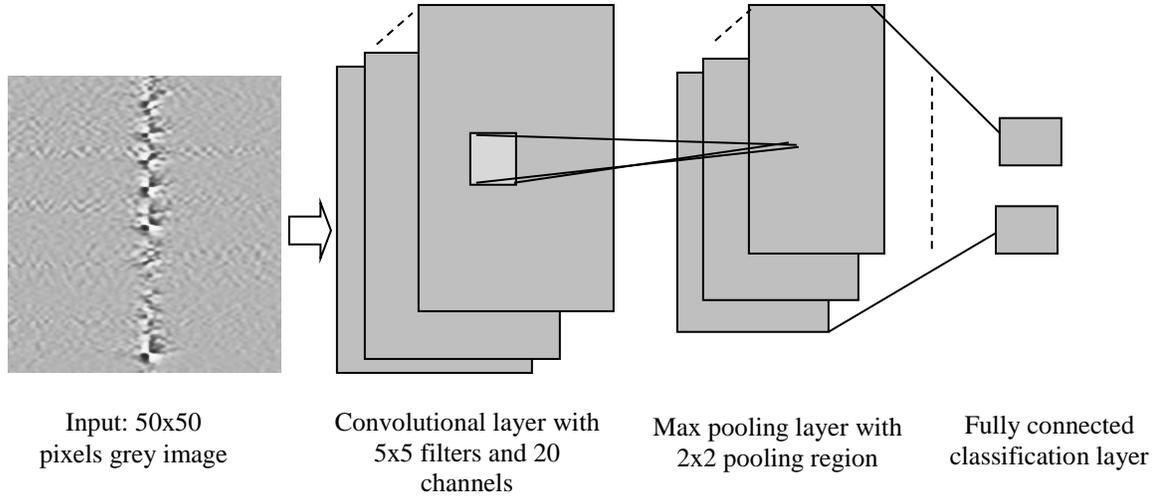

Input: 50x50 pixels grey image — Convolutional layer with 5x5 filters and 20 channels — Max pooling layer with 2x2 pooling region — Fully connected classification layer

Figure 7. Deep Learning Convolutional Neural Networks

The Deep Neural Networks (DNN) from Figure 7. consists of the following layers:
- 2D Convolutional Layer consisting of 5x5 filters with 20 channels (feature maps) in the output of the convolutional layer. The output of each filter x is passed to a rectified linear unit that produces max(0,x) as its output.
- 2D Max. Pooling Layer dividing the input into rectangular regions and returning the maximum value of each region. The height and width of the rectangular region (pool size) are both 2. This layer creates pooling regions of size 2x2 and returns the maximum of the four elements in each region. Because the stride (step size for moving along the images vertically and horizontally) is also 2x2, the pooling regions do not overlap.
- The images are classified into two classes (Excavator, no Excavator) by a fully connected output layer that uses SoftMax function that assign output $y_k$ for each output class k:

$$y_k = \frac{exp(\theta_k^T x)}{\sum_{j=1}^{K} exp(\theta_j^T x)} \qquad (1)$$

Where $\boldsymbol{\theta}_k$ are input parameters of the neuron that outputs the probability $y_k$ of the class k.

By training the network the parameters are adapted using stochastic gradient descent algorithms with the momentum update of the parameters $\boldsymbol{\theta}$ according to the following equations:

$$\boldsymbol{\theta} = \boldsymbol{\theta} - v \qquad (2)$$

Where

$$v = \gamma v + \alpha \nabla_\theta E(\boldsymbol{\theta}) \qquad (3)$$

And

$$E(\boldsymbol{\theta}) = - \sum_{i=1}^{n} \sum_{j=1}^{k} t_{ij} \ln y_j (x_i, \boldsymbol{\theta}) \qquad (4)$$

Where $\boldsymbol{\theta}$ is the parameter vector, $t_{ij}$ is the indicator that the *i*th sample belongs to the *j*th class, and $y_j(x_i, \boldsymbol{\theta})$ is the output for sample *i*. The output $y_j(x_i, \boldsymbol{\theta})$ can be interpreted as the probability that the network associates *i*th input with class *j*, i.e., P(t_j=1|x_i).

The network is trained with maximal 50 epochs with random batches of size n = 128 (default value), initial learning rate $\boldsymbol{\alpha}$ of 0.001 and momentum constant γ set to 0.9 (default value).

In contrast to the classic ML approach, where each second data should be evaluated and possible detected events tracked over several seconds up to few minutes, our imaging based approach don't need tracking, since each image already encompass several tens of seconds up to few minutes i.e. contains already a track of an event. Furthermore, the DNN outputs event probabilities without the need for the probability evaluation of single detections over the whole track like in the case of classic ML approach.

The performance of DNN in comparison to Classic ML approach described above is depicted in Table 2.

**Table 2: Comparison of classic ML approach with image based DNN approach**

|  | Minimum delay for one false alarm per month [s] | Execution time (17 km fibre, 60s data) [s] | Max detection distance [m] |
|---|---|---|---|
| Classic ML | 90 | 60 | 30 |
| Image + DNN | 15 | 5 | 10 |

As can be seen from Table 2 image and DNN based approach outperforms classic ML approach regarding both minimum delay required for reliable detection and execution time i.e. needs six times lower delay and twelve times lower execution time. Only the maximum distance where the reliable detection can be achieved is higher in the case of classic ML based approach.

## 5. CONCLUSIONS AND FURTHER WORK

As shown above, using images of the sensor signals and deep neural networks for pattern recognition is a promising approach for event detection with distributed acoustic sensing. We could achieve better recognition performance i.e. lower delay and execution time than with classic machine learning approach. Furthermore, we do not need explicit feature selection procedures like in classic machine learning approach that require high implementation effort and good domain knowledge.

In this work, we used off the shelf deep learning convolutional neural network without much hyperparameter optimization. In future, further optimization of the deep neural network can be done that enables detections of an excavator also on the larger distances as well as detection of some other interesting events like manual digging or welding. Transfer learning by usage of pre-trained neural networks is also an interesting topic for further research.

The method used in this work: Converting sensor signals into images and then use deep neural networks for pattern recognition in images, can be also used for event detection in other multi-sensor systems. We think it is a promising approach for optimal sensor fusion in multi-sensor systems.

## 6. ACKNOWLEDGMENT

I would like to thank my colleagues from WS Technology: Thomas Berger, Markus Dorn, Thomas Fröhlich, Albrecht Grabner, Richard Kindermann, Christian Magerle, Willhelm Plotz, Roman Steinwendtner, Kuno Zhuber-Okrog. Without them this work could not be possible.

# REFERENCES

[1] Da-Peng Zhou, Zengguang Qin, Wenhai Li, Liang Chen, and Xiaoyi Bao, Distributed vibration sensing with time-resolved opticalfrequency-domain reflectometry , Vol. 20, No. 12 / OPTICS EXPRESS p.13138J.P, June 2012 .

[2] Juarez at. al., 2005, Distributed Fiber-Optic Intrusion Sensor System , Journal Of Light Wave Technology, Vol.23, No.6, June 2005.

[3] J. A. Bucaro and T. R. Hickman, Measurement of sensitivity of optical fibers for acoustic detection, Applied Optics, Vol. 18, Issue 6, pp. 938-940, 1979.

[4] Charles M. Davis, Fiber Optic Sensors: An Overview, Opt. Eng. 24(2), April, 1985.

[5] A. J. Rogers, Distributed optical-fibre sensors for the measurement of pressure, strain and temperature, Physics Reports, Volume 169, Issue 2, Pages 99–143, October 1988.

[6] J. N. Fields, C. K. Asawa, O. G. Ramer and M. K. Barnoski, Fiber optic pressure sensor, J. Acoust. Soc. Am. 67, 816, 1980.

[7] Kyoo Nam Choi, Juan Carlos Juarez, Henry F. Taylor., Distributed fiber-optic pressure/seismic sensor for low-cost monitoring of long perimeters , Proceedings of SPIE, Vol. 5090, 2003.

[8] Tatsuya Kumagai, Shinobu Sato and Teruyuki Nakamura, Fiber-Optic Vibration Sensor for Physical Security System , IEEE International Conference on Condition Monitoring and Diagnosis, September 2012.

[9] Henry F. Taylor and Chung E. Lee, Apparatus and Method for Fiber Optic Intrusion Sensing, US Pat. US5,194,847 Granted March 16, 199.

[10] Harman, Robert Keith Fiber Optic Interferometric Perimeter Security Apparatus and Method, Int. Patent Application WO2013/185208, Publication Date December 19 2013.

[11] R. Allen, D. Mills, "Signal analysis: Time, Frequency, Scale, and Structure", Wiley, New York, 2004.

[12] R. Duda, P. Hart, David G. Stork, "*Pattern Classification*", 2nd Edition, Wiley, 2000.

[13] S. Theodoridis, K. Koutroumbas, "*Pattern Recognition*", 4th Edition, Academic Press, 2008.

[14] A. Krizhevsk., I. Sutskever, G. Hinton, ImageNet classification with deep convolutional neural networks. In Proc. Advances in Neural Information Processing Systems 25 1090–1098 2012.

[15] I. Goodfellow, Y. Bengio, A. Courville, "*Deep Learning*", MIT Press, 2016.

[16] S. T. Roweis and Z. Ghahramani, "A unifying review of linear Gaussian models", Neural Computation", 11(2):305–345, 1999.

[17] S. Arulampalam, S. Maskell, N. Gordon, and T. Clapp, "A Tutorial on Particle Filters for On-Line Non-Linear/Non-Gaussian Bayesian Tracking", IEEE Trans. Signal Processing, vol. 50, no. 2, pp. 174-189, 2002.

[18] R.C. Gonzalez, R. E. Woods, "*Digital Image Processing*", 3rd edition, Prentice Hall 2007.

[19] V. Hlavac, M. Sonka, R. Boyle, "Image Processing, Analysis and Machine Vision", 4th edition, Cengage Learning, 2014.

[20] J. Pearl, "*Probabilistic Reasoning in Intelligent Systems*", Morgan Kaufmann, 1988.

[21] R.D Neapolitan, "Learning Bayesian Networks", Prentice Hall, 2004.

# AUTHORS' BACKGROUND

| Your Name | Title* | Research Field | Personal website |
|---|---|---|---|
| Mugdim Bubln | Dr.techn. | Machine Learning | |
| | | | |
| | | | |
| | | | |

*This form helps us to understand your paper better, the form itself will not be published.

*Title can be chosen from: master student, Phd candidate, assistant professor, lecture, senior lecture, associate professor, full professor